\pdfoutput=1

\documentclass[11pt]{article}

\usepackage{EMNLP2022}

\usepackage{times}
\usepackage{latexsym}
\usepackage{color}

\usepackage[T1]{fontenc}

\usepackage[utf8]{inputenc}

\usepackage{microtype}

\usepackage{inconsolata}

\usepackage{amssymb}
\usepackage{pifont}
\newcommand{\cmark}{\ding{51}}%
\newcommand{\xmark}{\ding{55}}%
\usepackage{makecell}
\usepackage{multirow}
\usepackage{graphicx}

\title{Symptom Identification for Interpretable Detection of Multiple Mental Disorders}

\author{Zhiling Zhang\thanks{~~Equal Contribution} \and Siyuan Chen$^*$ \and Mengyue Wu\thanks{~~Corresponding Authors} \and Kenny Q. Zhu$^\dagger$\\
Shanghai Jiao Tong University \\
\texttt{\{blmoistawinde, chensiyuan925, mengyuewu\}@sjtu.edu.cn, kzhu@cs.sjtu.edu.cn} }

\begin{document}
\maketitle
\begin{abstract}
Mental disease detection (MDD) from social media has suffered from poor generalizability and 
interpretability, due to lack of symptom modeling.
This paper introduces \textbf{PsySym}, the first annotated symptom identification corpus 
of multiple psychiatric disorders, to facilitate further research progress. 
  PsySym is annotated according to a knowledge graph of the 38 symptom classes related to 7 mental diseases complied from established clinical manuals and scales, and a novel annotation framework for diversity and quality. 
  Experiments show that symptom-assisted MDD enabled by PsySym can outperform strong pure-text baselines. We also exhibit the convincing MDD explanations provided by symptom predictions with case studies, and point to their further potential applications. 
  \footnote{Code and dataset can be provided upon request.}
\end{abstract}

\section{Introduction}

Mental health has been a significant challenge in global healthcare. Nearly 1 in 5 U.S. adults live 
with a mental illness or condition~\citep{NIMHMental}, and there are about 1 billion people suffering from mental disorders worldwide~\citep{UNMental}. Due to the stigma of mental disorders and lack of professional mental health services, many people cannot receive proper diagnose or treatment for their conditions. Social Media can be a promising source for studies on this problem, as we may detect hidden traces of mental disorders from the symptoms that users may reveal in their free sharing, and provide pertinent help for those in need. Consequently, Mental Disease Detection (MDD) from social media has received increasing attention \citep{coppersmith2015adhd,cohan2018smhd}. 

\begin{figure}[t]
    \centering
    \includegraphics[width=1.0\columnwidth]{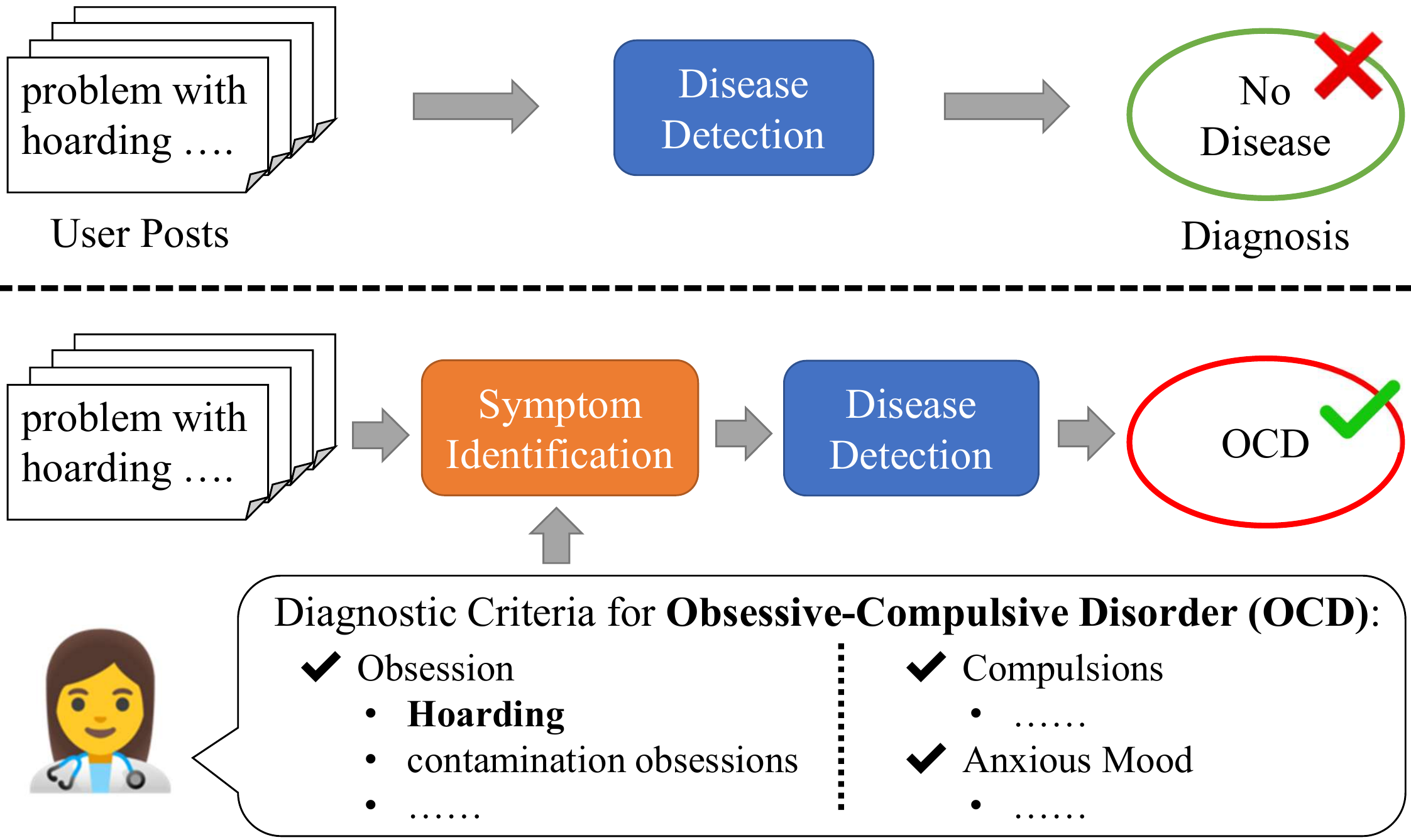}
    \caption{Comparison between text-based and the proposed symptom-assisted 
mental disease detection method, which leverages psychiatric knowledge for 
symptom identification to improve the effectiveness and interpretability of MDD.}
    \label{fig:overview}
\end{figure}

However, most automatic MDD methods still struggle in this task, especially for their unsatisfying generalizability and explainability. Firstly, these models may learn dataset-specific spurious correlations between certain words and the labels (usually diseases), and thus fail to generalize~\citep{harrigian2020models}. Moreover, most deep learning based methods work as black boxes, and cannot provide explanations for their prediction, which differs from the clinical practice which leverages symptom-based diagnostic criterions from authoritative manuals like DSM-5 \citep{american2013diagnostic}. Therefore, it may be hard for current MDD methods to gain trust from their users. 

To tackle these issues, there has been a rising interest in utilizing \textit{symptoms} for MDD, as they are the bases that human psychiatrists use to make diagnoses. Pioneering research has shown their potential benefits of improving the accuracy, generalizability and interpretability of MDD \citep{lee2021micromodels,nguyen2022improving,zhang2022psychiatric}. Nevertheless, due to the lack of large-scale annotated corpus for supervised learning, they can only extract symptom features with unsupervised/weakly supervised methods or simple pattern matching, which may not guarantee the quality of the extracted features. Moreover, most of these works only focus on the detection of depression. However, many mental disorders may share similar set of symptoms. For example, depressed mood can not only be seen on the patient of depression, but also those suffering from bipolar disorder, anxiety, etc. Jointly modeling the symptoms of multiple diseases may enhance the performance on all classes. 

These limitations call for the establishment of a large-scale, multi-disease annotated dataset for symptom identification, which will face many novel challenges. Initially, the symptoms of multiple diseases are scattered over the different chapters of DSM-5 and other materials. Similar symptoms would have varied expressions in different places, which causes difficulty in setting up the annotation standard. Furthermore, the free and diverse language style on social media \citep{yadav2020identifying} can make the retrieval of candidate posts for annotation difficult. Last but not least, the relatively large amount of symptoms from different disorders and the nuanced differences and similarities between them make it hard to get high-quality annotations (i.e. inter-rater agreement will be low).  

In this work, we propose a novel data annotation framework, and introduce the first multi-disease symptom identification dataset based on social media posts, \textit{PsySym} (\textbf{Psy}chiatric-disorder \textbf{Sym}ptoms), which contains the multi-label annotations of 38 symptom classes from 7 mental diseases on 8,554 Reddit post sentences. We establish our annotation target (symptom classes) mainly based on the diagnostic criterions from DSM-5, with symptom descriptions on clinical questionnaires as supplementary. We leverage embedding-based retrieval methods \citep{reimers-2019-sentence-bert} instead of keyword matching to get the candidate sentences for annotation, which can effectively constrain our efforts to a precise but diverse subset of posts for efficient annotation. To guarantee the quality of data, we apply several quality control approaches, and divide the annotation tasks by separate diseases so as to reduce the cognitive burdens of annotators, resulting in high inter-rater agreement.

Finally, we propose a symptom-assisted MDD framework (Figure \ref{fig:overview}). We use models trained on PsySym to extract symptom features for MDD, outperforming strong BERT-based baseline \citep{devlin2018bert}. We also explore the interpretability of symptom identification for MDD. We find that they can reasonably provide DSM-5 compliant explanations for diagnosed patients. We will also show that symptom-based interpretations can also help us find incorrect labels in automatically constructed MDD datasets, indicating their further potential. 

Our contributions are:
\begin{itemize}
    \item We build the first social-media based symptom identification dataset of multiple mental diseases, \textit{PsySym}, with novel annotation framework to guarantee the diversity and quality of the dataset.
    \item We propose symptom-assisted MDD, which leverages the features extracted from PsySym-trained models, and can outperform strong baselines in MDD.
    \item We demonstrate the intuitive interpretability for MDD results enabled by symptoms, and its promising applications with case studies. 
\end{itemize}

\section{Related Work}

\paragraph{Mental Disease Detection} Mental Disease Detection (MDD) from social media is enabled by the users' self disclosure of their diagnosis, or their participation in mental-disease related topics and forums. These proxy signals can be leveraged to automatically label the diagnosed diseases of users for the supervised learning of machine learning algorithms. Early researches mainly focus on the detection of depression \citep{de2013predicting}, and following works further extend the scope to multiple diseases \citep{coppersmith2015adhd,cohan2018smhd}. 

Approaches for MDD can be mainly divided into two types. The first utilizes features like bag-of-words, topic modeling and LIWC \citep{pennebaker2001linguistic} with traditional machine learning algorithms \citep{shen2017depression,trotzek2018utilizing}. These methods can provide word/topic-level interpretability, but the cannot leverage the temporal pattern of the posts. The second type leverage deep neural networks that can encode the posts as a sequence for better temporal modeling \citep{yates2017depression,sekulic2019adapting,gui2019cooperative}. However, these methods work as black-boxes and cannot provide explanations. Recent works have also revealed that both types of methods suffer from the lack of generalizability \citep{harrigian2020models,nguyen2022improving}. 

\paragraph{Symptom Identification} There have been some pioneering attempts to leverage symptom-related features for MDD. \citet{karmen2015screening} uses manually complied lexicon to detect symptoms, and aggregate them into a score for the detection of depression. \citet{lee2021micromodels} and \citet{zhang2022psychiatric} leverage the embedding similarity between a post and symptom-related descriptions to decide the presence or risk of a symptom. \citet{nguyen2022improving} uses regular expressions and heuristics to automatically build weakly-supervised training data for symptom identification. These methods have exhibited superior generalizability and interpretability. Nevertheless, the efforts on establishing annotated corpus for symptom identification are still limited \citep{mowery2017understanding}, which may hinder the potential of symptom-assisted MDD methods for leveraging stronger supervised symptom models. 

\section{Dataset Construction}
In this section, we introduce the construction of \textit{PsySym}, the first annotated multi-disease symptom identification dataset based on social media posts.

\subsection{Disease-Symptom Knowledge Graph}
\label{sec:kg}

\begin{figure}[ht]
    \centering
    \includegraphics[width=1.0\columnwidth]{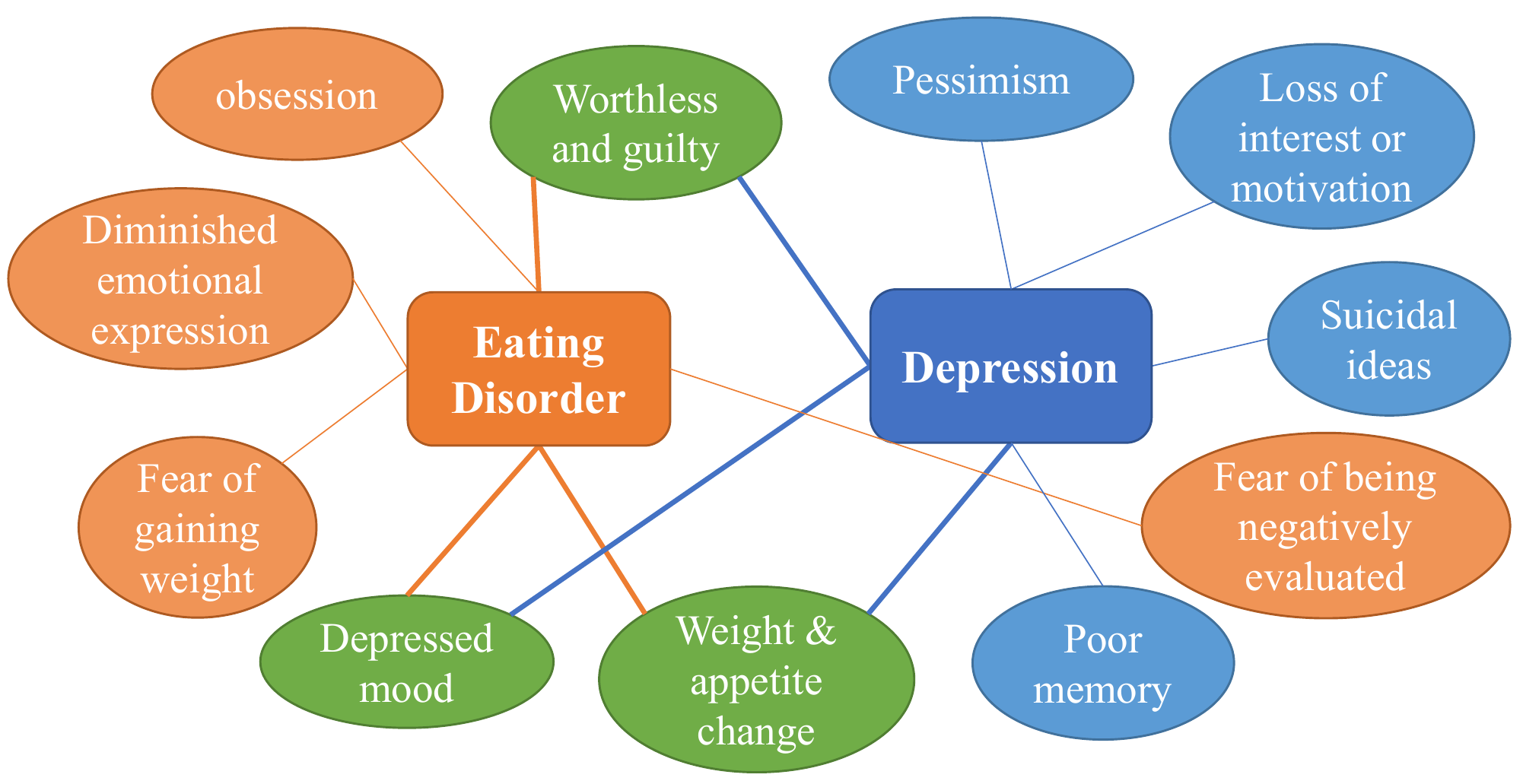}
    \caption{Illustration of the established disease-symptom knowledge graph. Only 2 diseases and their corresponding symptoms are shown for clarity. Many symptoms can be shared by multiple diseases, like the green nodes in the figure.}
    \label{fig:symp_kg}
\end{figure}

Before data construction, we need to decide the annotation targets, i.e. which diseases and their corresponding symptoms to annotate. Considering our downstream application of MDD, we choose 7 diseases\footnote{We initially try to annotate the symptoms of all 9 diseases of SMHD, but we find it hard to plausible samples for \textit{Autism} and \textit{Schizophrenia}, and thus focus on the remaining 7.} that are used in the established SMHD dataset \citep{cohan2018smhd}: \textit{Depression, Anxiety, ADHD, Bipolar Disorder, OCD, PTSD, Eating Disorder}. We then find symptoms used in the diagnostic criterion of these diseases within DSM-5~\citep{american2013diagnostic}. However, the symptoms in DSM-5 are not represented as standard classes, but expressed in natural language with similar symptom having nuanced differences in different places. Therefore, we manually merge similar expressions of a symptom into one standardized class, and store the expressions as its detailed descriptions (also referred to as \textit{sub-symptoms}). After merging the symptoms, the diseases and the standardized symptoms constitute a bipartite Knowledge Graph (KG), where we can clearly see the shared symptoms of different diseases. We also supplement the KG with representative clinical questionnaires, where we also merge the symptoms mentioned in each question/item into the standard classes, and add links between the targeted disease and the measured symptoms if such edges are not found in DSM-5. The final knowledge graph (Figure \ref{fig:symp_kg}) has 45 nodes (7 diseases and 38 symptoms), and 162 edges. Symptoms like \textit{Depressed Mood} and \textit{Inattention} are shared by as many as 5 diseases, suggesting the potential of learning such shared features with multi-disease modeling. 

\subsection{Annotation Candidates Retrieval}
\label{sec:data_retrieval}

\begin{table*}[t]
    \small
    \centering
    \begin{tabular}{|l|l|l|}
    \hline
        Post & Symptom(s) & Status  \\ \hline
        Libido, pls come back! & Genitourinary (loss of libido) & True  \\ \hline
        I feel sad and motivationless. & Depressed; Loss of Interest or Motivation & True  \\ \hline
        I'm questioned if I'm manic. & Mood Shift & Uncertain  \\ \hline
    \end{tabular}
    \caption{Example annotations in PsySym. We can match the first post with a sub-symptom (loss of libido) of genitourinary system despite its figurative style. The second post has multiple labels. The symptom in third post is ambiguous, so we provide a distinct ``Uncertain'' label for its symptom status.}
    \label{tab:label_example}
\end{table*}

We then search for candidate posts to annotate the symptoms. We choose Reddit as our data source for its public availability and wide acceptance in previous literature \citep{losada2016test,cohan2018smhd,wolohan2018detecting}. Specifically, the candidate pool consists of all self-posts from 2005 to 2016 in the PushShift dataset \citep{baumgartner2020pushshift}, and all posts are split into sentences for later usage. 

The huge amount of posts necessitate a pre-step of selecting candidates that are likely to express certain symptoms for acceptable annotation efforts. First, we only select candidates from mental health related subreddits \citep{cohan2018smhd}, where more posts will be relevant. Moreover, we leverage embedding-based retrieval to further narrow down the range\footnote{We designed an algorithm to promote the balance of all symptoms classes. Details in Appendix \ref{apd:cand}.}. Specifically, we use Sentence-BERT \citep{reimers-2019-sentence-bert} to encode the post sentences and the symptom descriptions in the KG into embeddings. Then we will calculate the cosine similarity between them, and estimate a sentence's relevance to a symptom with its max similarity with all of the symptom's sub-symptoms. The rationale is that a symptom can have different manifestations expressed in its descriptions, and a sentence can be considered to convey that symptom if it resembles any one of them. However, since the original symptom descriptions in DSM-5 are expressed in a professional style and usually observed as a third party, they do not necessarily reflect 
the self-reporting nature of the content on social media. 
The descriptions collected from clinical questionnaires, however, can alleviate the problem, as they are designed to be easy to understand and fill in. To further tackle the mismatching that cannot be solved even with the questionnaires, we also directly collected some typical Reddit posts about a symptom for the similarity calculation in place of the official descriptions. \footnote{We provide experimental results in Appendix \ref{apd:exp_retrieve} showing that our final method can lead to better precision and recall for the retrieved candidates, compared with the keyword/pattern matching methods commonly used in previous works \citep{mowery2017understanding,yadav2020identifying}.}

\subsection{Annotation Design}
\label{sec:data_annotation}

Annotations for symptom identification usually involve the binary decisions if a symptom can be identified from the sentence. Such decision can sometimes be tricky when a symptom is mentioned in a sentence, but is not actually present. For example, ``I don't have panic attack.'' implies a \textit{negation} of the symptom, and ``Is it panic attack?'' expresses the \textit{uncertainty} about the symptom. Although most previous works like \citet{nguyen2022improving} treat such cases as negative samples, we think that they are different from the other negatives like sentences totally irrelevant to any symptoms or only about other symptoms. Distinguishing between them may enable a more informative analysis and benefit downstream applications. Therefore, we decided to divide the annotation into two tasks: \textbf{relevance judgment} and \textbf{status inference}, as are exemplified in Table \ref{tab:label_example}. For relevance judgment, the annotator needs to judge whether the sentence is relevant to the given symptoms. Note that the symptoms can be described in figurative language instead of standard patterns, and they can be negated or uncertain. For status inference, the annotator needs to decide, if the relevant symptom(s) are indeed present. We denote positive/negative status as `True'/`Uncertain'.

Before crowdsourcing annotation, we first conducted preliminary annotations ourselves. We found it hard to annotate all 38 symptoms among the candidates from all mental health subreddits. The relatively large number of classes, and the nuanced differences and similarities between them can pose heavy cognitive burden to the annotators. Therefore, we decided to separate the annotation job queues by disease. In each queue, the annotators only need to read the posts from the subreddits of that disease, and the symptoms are restricted to only the typical symptoms of the disease according to our KG. Although this transformation can potentially affect the recall of atypical symptoms, we find it significantly improve the annotation efficiency and agreement in our preliminary tests. 

We then invite volunteers for the annotation tasks, who are all well-educated and also include professional psychiatrists. To ensure the data quality, each sentence is annotated by 3 participants, and we also utilize a series of quality control protocols. The annotation proceeds as follows: 
\begin{enumerate}
    \item Training Session: We train the annotators about the annotation rules and also demonstrate some example annotations by ourselves through video meetings.
    \item Screening Tests: We collect test questions from samples on which the authors have consensus in preliminary annotations. The invited volunteers need to first annotate on these questions and achieve certain score to be eligible for further annotation. They can take the test for several times, and we require them to read the reason for our decision after the test for their better alignment with our requirements.
    \item Annotation: Those who passed the screening tests can proceed for further annotation. During annotation, they can always discuss with us about any questions encountered.
    \item Sampling Inspection: At intervals, we will sample 10\% of the completed annotations from an annotator for checking. We will correct any annotations we find inappropriate, and give a score according to the number of corrections. If the score is below certain threshold, all annotations in the checking batch will be rejected for re-labeling.
\end{enumerate}

Finally, we recruited 31 volunteers contributed valid annotations for 8,554 sentences. The average Fleiss's $\kappa$ for the relevance judgement of all 38 symptoms is 0.7708, and the $\kappa$ of status inference is 0.2518. Among all symptoms, \textit{Anxious Mood} has the most labels (1764), while \textit{Avoid Stimuli} has the least (78). For status inference, the `Uncertain' annotation constitutes 13.75\% for all annotations. More details are provided in Appendix \ref{apd:stats}.

\subsection{Labels and Data Splits}

To merge the multiple annotations into a single gold label for each sentence, we consider a symptom to be relevant to a sentence as any one of the annotation is positive, and we use the portion of \textit{uncertain} annotations as the label for status inference (more discussion in \S \ref{sec:model_status}). We split the dataset into training/validation/testing set by 5:1:4 to preserve enough samples for all classes in the test set for a stable evaluation. 

To allow models to accurately identify symptoms from all posts of social media, where the majority of them are not related to any mental disease symptoms, we also collect such posts (referred to as \textit{control posts}) for PsySym from the same candidate pool. We randomly sample posts whose author doesn't have any post or comment in mental health related subreddits, and further remove posts that contain any mental health related terms provided by \citet{cohan2018smhd}. Finally, we randomly sample sentences from the remaining posts, resulting in 83,779 sentences, distributed into training/validation/testing set by 5:1:4.

\subsection{Disease Detection Dataset}
\label{sec:data_disease}

To demonstrate the helpfulness of PsySym for the downstream task of MDD, we also construct a dataset by reimplementing the data collection method of SMHD \citep{cohan2018smhd}. We find diagnosed users by the pattern-matching of two components: one that indicates a self-reported diagnosis (e.g. ``diagnosed with'), and another that maps relevant keywords to the 9 mental diseases (e.g. ``panic disorder'' to \textit{Anxiety}). A user is labeled with a disease if one of its keywords occurs within 40 characters of the diagnosis pattern. Control users are randomly sampled from those who never posted or commented in mental health related subreddits. Similar to SMHD, we eliminate the diagnostic posts from the dataset to prevent the direct leakage of label, but we don't remove other mental health related posts to allow the extraction of symptom-related features. The final dataset consists of 5,624 diagnosed users and 20,981 control users with average number of posts per user being 102.5 and 119.4, respectively. We provide the distribution of each disease in Appendix \ref{apd:stats}. 

\section{Models}

This section will introduce the proposed models for the two sub-tasks of symptom identification: \textit{relevance judgment} and \textit{status inference} that can leverage the data from multiple diseases simultaneously, and how to leverage the above models for \textit{mental disease detection}.

\subsection{Symptom Relevance Judgment}
\label{sec:model_rel}
The task of relevance judgment can be viewed as a multi-label classification problem, where we need to predict for each symptom if it is relevant to the sentence. However, when we want to train the model on the annotations from different diseases, we will face the problem of \textit{missing labels}, as we will not annotate the relevance of symptom $s$ in the dataset of disease $d$ if $s$ is not considered to be a typical symptom of $d$. A naive solution is to treat all such missing labels as negative. However, since the 
co-existence of multiple diseases on the same person is not uncommon, it is likely that the 
typical symptom of other diseases will also be present. Therefore, 
such solution will lead to false negatives and harm the model performance. 

Inspired by \citet{fonseca2020addressing} and \citet{gururani2021semi}, we experimented with two techniques to address the problem of missing labels.

\paragraph{Loss Masking} Missing labels will be ignored during the loss calculation. This can prevent the model from learning incorrect negative labels, but also restricted the exploitation of the true negatives among missing labels.

\paragraph{Label Enhancement} This method involves two-stage trainings of a teacher model and a student model. First, a teacher model is trained using \textit{Loss Masking}. Then we use the teacher model to predict the probabilities of the missing labels in the training set. If the predicted probability of a symptom is lower than certain threshold, we change its label to negative, otherwise it will still be treated as missing. A second student model will be trained on the enhanced labels together with \textit{Loss Masking}. In contrast to previous works which set the threshold to enhance the missing labels with top $k\%$ confidence, we search for the cutting point where the teacher model can achieve 90\% True Negative Rate on the ROC curve of existing annotations as the threshold for each symptom, which can quantitatively guarantee the quality of the enhanced labels.

Overall, we use a BERT-based encoder \citep{devlin2018bert} with a linear layer on top of the representation of \texttt{[CLS]} to predict the probabilities of all symptoms
with a sigmoid activation, and train the model with binary cross entropy loss against the labels adjusted with the methods above.

When trained on PsySym with control posts, the especially unbalanced positive/negative ratio can make the training hard. We thus additionally implement a \textit{Balanced Sampler} which samples equal amount of annotated sentences and control sentences for each batch.

\subsection{Symptom Status Inference}
\label{sec:model_status}
Status Inference aims to predict if the symptoms relevant to the sentence are truly present instead of being a negation (e.g. denying or recovery) or an uncertain guess. Therefore, it is natural to formulate it as a single-label binary classification problem. However, due to the ambiguity in the expression,
lack of context and the different understandings of annotators, the agreement on the status labels are relatively low (\S \ref{sec:data_annotation}) despite our quality control efforts. Consequently, even if we can derive binary labels from the majority voting of annotators, models trained on such ambiguous targets will hardly show satisfying performance. 

However, we may not simply attribute such disagreement to poor annotation quality, since there is inherent ambiguity in the annotations of natural language inference tasks, as is reported by \citet{nie2020can}. We can still make reasonable probabilistic estimation of the status by embracing the ambiguity and directly learn from the annotation distribution \citep{meissner2021embracing}. Therefore, we change the learning target from binary labels to the portion of annotators who label the status as \textit{uncertain}, and the possible values are thus 0, 1/3, 2/3 and 1. Then we use BERT-based model with sigmoid activation and cross entropy loss to predict the non-binary labels. 

\subsection{Mental Disease Detection}
\label{sec:disease_detect}

\begin{figure}[h]
    \centering
    \includegraphics[width=1.0\columnwidth]{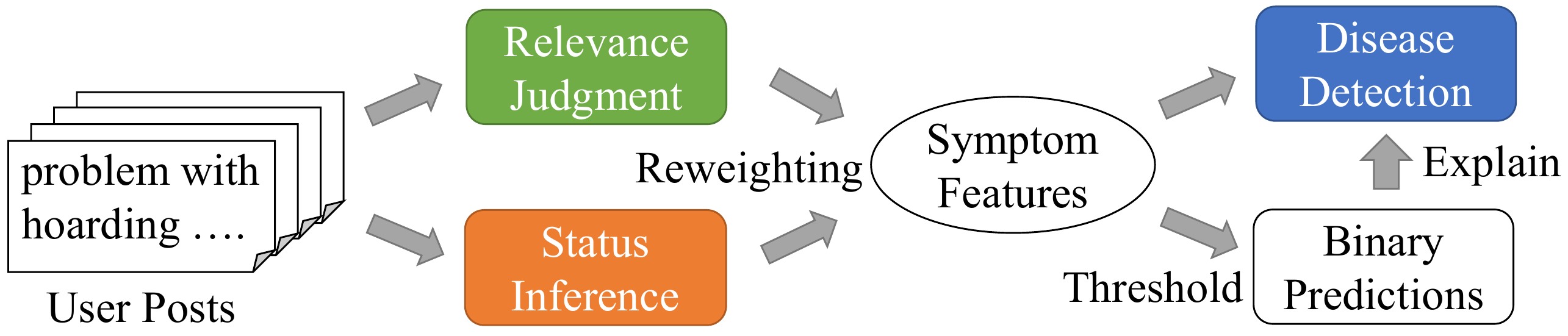}
    \caption{The proposed symptom-assisted MDD pipeline.}
    \label{fig:mdd_pipeline}
\end{figure}

The task of MDD is to predict if a user suffers from certain mental diseases with his/her posting history. Figure \ref{fig:mdd_pipeline} illustrates the proposed symptom-assisted MDD pipeline. First, we utilize the predicted probabilities from relevance judgment models trained on PsySym as the extracted symptom features. 

Next, to further improve the symptom features, we also introduce \textit{subject} and \textit{status} feature. The \textit{subject} feature is a binary variable indicating if the discussed symptoms of a post is about the poster himself. We calculate this feature by counting the mentions/pronouns of other people and the use of first person pronouns, and set the feature as 1 if the latter is no less than the former. The \textit{status} feature is the predicted probability of the status inference model that the symptoms are present. It is obvious that a post not about the poster himself or exhibiting symptoms clearly should not count much to disease detection. We thus experiment with the \textbf{Reweighting} approach similar to \citet{karmen2015screening} to incorporate them into the symptom features:
\begin{equation}
    f_{symp} = p_{rel} \times w_{status} \times w_{subj}
\end{equation}
where $p_{rel}$ is the probabilities predicted by the relevance model; $w_{status}$ is the probability predicted by the status model; $w_{subj} = 0.9$ if the subject is the poster, otherwise $w_{subj} = 0.1$.

To conduct MDD, we incorporate these features into the model proposed by \citet{nguyen2022improving}. This model utilizes CNN of various kernel sizes on top of the sequence of feature vectors extracted from a user's posting list to aggregate the information from consecutive posts. The features can either be pure-text features like the sentence embeddings from pretrained BERT, or the proposed symptom features (denoted as \textbf{Symp} below). Note that the symptom features can be much more condense than pure-text features (38-dim symptom probabilities versus 768-dim BERT embedding).

Finally, for the ease of explaining MDD results with symptoms (\S \ref{sec:interpret}), we may need binary decision on the their presence. To achieve this, we use 0.5 to threshold on the reweighted symptom features, where the re-weighting procedure can help eliminate posts that should not be counted for the diagnose. 

\section{Experiments}
\label{sec:exp}

In this section, we present experimental results to: (1) exhibit the benefits brought by PsySym's design choices such as multi-disease modeling, and the incorporation of status inference for symptom identification. (2) examine the effectiveness of symptom features for MDD. (3) demonstrate the interpretability enabled by symptom identification for MDD.

\subsection{Methods of Comparison}

For all prediction tasks, we mainly compared the proposed methods with 2 types of baselines: \textbf{TF-IDF+LR} is a representative non-deep learning method which utilizes TF-IDF to extract textual features, followed by a Logistic Regression model for prediction. \textbf{BERT/MBERT} uses pretrained, \texttt{base} size of BERT and MentalBERT \citep{ji2021mentalbert}, which can establish a strong baseline. More details like hyperparameter settings can be seen in Appendix \ref{apd:settings}.

\subsection{Symptom Relevance Judgment}

For symptom relevance judgment, we first conduct experiments on PsySym without control posts mainly to check the effectiveness of different modeling choices. We report the performance in Table \ref{tab:symp} according to the threshold-free metric AUC, averaged across each symptom class in the subset of each disease.

\begin{table}[h]
    \small
    \centering
    \begin{tabular}{clc}
    \hline
    \multicolumn{1}{l}{} & Method & AUC   \\ \hline
                         & TF-IDF+LR                      & 87.86 \\
                         & BERT                           & 91.60 \\
    \multirow{-3}{*}{\begin{tabular}[c]{@{}c@{}}Single\\ Disease\\ (7 models)\end{tabular}} & MBERT                 & \textbf{91.77} \\ \hline
                         & MBERT                          & 91.00 \\
                         & MBERT (loss mask)              & 92.21 \\
    \multirow{-3}{*}{\begin{tabular}[c]{@{}c@{}}Multi\\ Disease\\ (1 model)\end{tabular}}   & MBERT (label enhance) & \textbf{92.94} \\ \hline
    \end{tabular}
    \caption{Symptom relevance judgment results on PsySym without control posts.}
    \label{tab:symp}
\end{table}

The single disease methods on the first 3 rows leverage models trained separately on the each disease subset. We can see that BERT significantly outperforms TF-IDF+LR, while the further pretraining on mental-health corpus done by MBERT can bring additional improvement. We thus use MBERT in the following experiments. The last 3 multi-disease methods only train one model on the combined dataset of all diseases, where we will tackle the problem of missing labels with the techniques introduced in \S \ref{sec:model_rel}. Comparing the third and fourth row, we can see that the multi-disease model's performance drops with the default strategy of treating all missing labels as negative. However, with \textit{Loss Masking}, the multi-disease model can now outperform the single disease counterpart, and \textit{Label Enhancement} brings additional gain. This proves our hypothesis that the simultaneous modeling of multi-disease data can help improve the relevance judgment performance. 

In order to predict symptom features for general user posts, we then train a relevance model on PsySym with control posts, leveraging the additional balanced sampler (\S \ref{sec:model_rel}). Its AUC is 98.54 and F1 (with threshold 0.5) is 67.03, averaged across 38 symptoms on the full test set containing all diseases and control posts, while directly transfer the model not trained with control posts would lead to only 30.53 F1.

\begin{table*}[t]
    \centering
    \small
    \begin{tabular}{lc}
    \hline
    \multicolumn{1}{c}{Posts (paraphrased for anonymity)}   & Predicted Symptom   \\ \hline
    I have a problem with hoarding.                         & Obsession           \\
    Compulsive nail biting is my problem and I also bathe compulsively.       & Compulsion   \\
    I am under so much stress that not even bath can make the anxiety go away. & Anxious Mood \\ \hline
    \multicolumn{2}{c}{Typical OCD symptoms: Obsession \cmark~ Compulsion \cmark~ Anxious Mood \cmark} \\
    \hline
    \end{tabular}
    \caption{The predicted symptoms of some posts by an OCD patient, which covered all typical OCD symptoms and constituted a convincing explanation for the diagnose. }
    \label{tab:explain_tp}
\end{table*}

\begin{table*}[t]
    \centering
    \begin{minipage}{.48\textwidth}
      \centering
      \small
      \begin{tabular}{l|l}
        \hline
Dataset Label      & Autism, \textcolor{red}{Eating disorder, PTSD}  \\
Disease Prediction & Autism  \\
Predicted Symptom  & \begin{tabular}[c]{@{}l@{}}{[}Autism{]} social problems~\cmark \\ {[}EAT{]} appetite change~\xmark\\ {[}PTSD{]} fear of trauma~\xmark \end{tabular} \\
        \hline
      \end{tabular}
    \end{minipage}
    \begin{minipage}{.48\textwidth}
      \centering
      \small
      \begin{tabular}{l|l}
        \hline
        Dataset Label      & OCD  \\
        Disease Prediction & OCD, \textcolor{teal}{Anxiety}  \\
        Predicted Symptom  & \begin{tabular}[c]{@{}l@{}}{[}OCD{]} obsession~\cmark \\ {[}Anxiety{]} anxious mood~\cmark\\ ~~~~~~~~~~~~~~~~~social anxiety~\cmark\end{tabular} \\
        \hline    
    \end{tabular}
    \end{minipage}
    \caption{Examples where our disease prediction model corrects the inaccurate labels produced by automatic methods. Symptom-based explanations can help detect such labeling errors and justify correct predictions.}
    \label{tab:explain_fp_fn}
\end{table*}

\subsection{Symptom Status Inference}

Since the status inference model is trained with the non-binary targets of annotation distribution, we use Mean Absolute Error (MAE) as the evaluation metric. To get a better grounding for understanding the model performance, we establish a no-model baseline  as the performance lower bound, using the mean probability in the test set as the prediction for all samples. We also use the expected MAE of a single annotator to estimate the performance upper bound. We train a Mental-BERT based model for status inference, achieving 0.1360 MAE, compared to a lower bound of 0.1940 and an upper bound of 0.1172, which indicates a plausible performance.

\subsection{Mental Disease Detection}

We show MDD performance in Table \ref{tab:disease} and compare the performance of methods utilizing pure text and symptoms. The training and evaluation for each disease is conducted in a binary setting, where the model needs to distinguish the diagnosed users of that disease and the control users, so users with other diseases will not be involved. We then report the average F1 across all 9 diseases. 

\begin{table}[h]
    \small
    \centering
    \begin{tabular}{lc}
        \hline
        Method          & F1             \\
        \hline
        TF-IDF+LR       & 43.73          \\
        BERT \citep{nguyen2022improving}        & 51.46          \\
        \hline
        Symp            & 55.46          \\
        Symp (Reweighting) & \textbf{57.09} \\
        \hline
    \end{tabular}
    \caption{Mental Disease Detection Results, averaged across 9 diseases.}
    \label{tab:disease}
\end{table}

We can see that \textit{Symp} outperforms all pure-text methods including the strong BERT model, suggesting the usefulness of symptom features for MDD. The \textit{Reweight} method that incorporates \textit{status} and \textit{subject} feature into symptom feature can bring further improvement, indicating that these additional aspects can help properly decide symptom risks for better MDD. 

\subsection{Case Study on Interpretability}
\label{sec:interpret}

One of the major goals of symptom identification is to enable machine learning models to provide explanations for disease diagnoses just as human psychiatrists. The binarized prediction of symptoms of our model, and the disease-symptom relations from our knowledge graph (derived from clinical manuals) can help achieve this. We provide concrete examples below. 

Table \ref{tab:explain_tp} shows that, for a patient predicted to have OCD, the symptom model with reweighting (\S \ref{sec:disease_detect}) can find all typical OCD symptoms from his/her posting history, and thus justify the diagnose.

Being able to interpret symptoms can also help us spot spurious diagnosis. Here we use the model to examine the correctness of disease labels produced by pattern matching based automatic method (\S \ref{sec:data_disease}). Despite its careful design, it can still make both false-positive and false-negative errors, as has been reported by \citet{cohan2018smhd}. Such problematic labels can negatively affect the generalizability of the model trained on them \citep{ernala2019methodological}, but they can be hard to detect. 

Symptom-based explanations may provide an efficient way to detect these false labels. As is shown in Table \ref{tab:explain_fp_fn} (left), we can easily find and remove the falsely labeled diseases when there are few or no history of their corresponding symptoms. For the example at right, \textit{Anxiety} is missed by the labeling method, while the high prevalence of anxiety symptoms like anxious mood and social anxiety predicted from the user's post can indicate its presence. Therefore, the interpretability brought by symptoms can justify correct predictions and may further serve as reference for human correction of labels.   

\section{Conclusions}

In this work, we introduce \textit{PsySym}, the first annotated multi-disease symptom identification dataset based on social media posts. A novel annotation framework is proposed to guarantee the diversity and quality of the annotations. PsySym defined two sub-tasks of symptom identification: relevance judgment and status inference, to enable a more comprehensive analysis. Strong baselines are established on the two sub-tasks with multi-disease modeling techniques that can properly handle missing labels and distribution-targeted learning to deal with the ambiguity in status inference. With PsySym-trained models, symptom-assisted MDD method can outperform strong pure-text baselines. Qualitative examples also demonstrate the explainability enabled by symptom predictions for MDD. 
\section*{Limitations}
\label{sec:limitations}
There are some limitations to this study that could be addressed in future research. 

First, although we tried our best to improve the diversity of the annotated sentences with embedding-based retrieval methods that can find symptom expressions without standard keywords. There can still be blind points we can not cover, such as the posts outside mental health related subreddits, and those cannot be found due to the limitations of the retrieval model itself.

Moreover, there are some types of symptoms we are unable to annotate due to the characteristics of our data source. For example, hallucination (a symptom of schizophrenia) usually requires the observations from another person to be identified, and can hardly be found on Reddit where user mainly shares subjective experience. The fact that Reddit is dominated by adult users \citep{gjurkovic2021pandora} also prevents us from finding the typical symptoms of autism and ADHD among children. 

Last but not least, our dataset does not include useful signals from modalities other than text. For instance, the time pattern of posting may also reveal the symptom of insomnia, and the features of the user's ego centric network may show the troubles in his/her social relations \citep{de2013predicting}. The faces and colors of the posted image may also help identify depression \citep{gui2019cooperative}. If videos or sounds can be leveraged, the acoustic features of speech can help recognize the emotions like sadness, fear and anger \citep{Busso2008IEMOCAPIE} for better detection of mental diseases \citep{wu2022climate}. 
\section*{Ethics Statement}

\paragraph{Annotation} We pay the annotators a fair wage above the minimum requirement.
If workers have any questions or concerns, we will respond to them immediately. 
Since the content involves the expression of mental disease symptoms, we may expect negative effects on the annotators. Therefore, the annotators can freely take breaks or quit the task at anytime. 
We also interviewed some annotators about their feeling after annotation. They only reported slight discomfort at the time of reading sad or frightening posts due to empathy, and they found no long-term negative effects on them.

\paragraph{Application} Mental disease detection can be related to some sensitive topics, so we should be careful with its applications. First, since mental diseases like depression are still not well understood or even stigmatized in many regions, improper usage of MDD techniques may do harm to the users. Moreover, the precision and recall of the algorithm is far from prefect. It may make false/missing diagnoses which can prevent the user from getting proper treatment, but may still be an useful auxiliary tool for those who are unaware of their mental conditions or cannot access mental services. Therefore, the predictions of the model should be carefully re-examined by professionals for a confirmed diagnosis, where the symptom prediction results may facilitate quick inspection when served as the disease-specific summary of the long posting history. 
We will also require the users of PsySym to comply with a data usage agreement to prevent the invasion of privacy or other potential misuses. 


\bibliography{anthology,custom}
\bibliographystyle{acl_natbib}

\appendix

\section{Data Construction Details}

\begin{figure*}[ht]
    \centering
    \includegraphics[width=1.8\columnwidth]{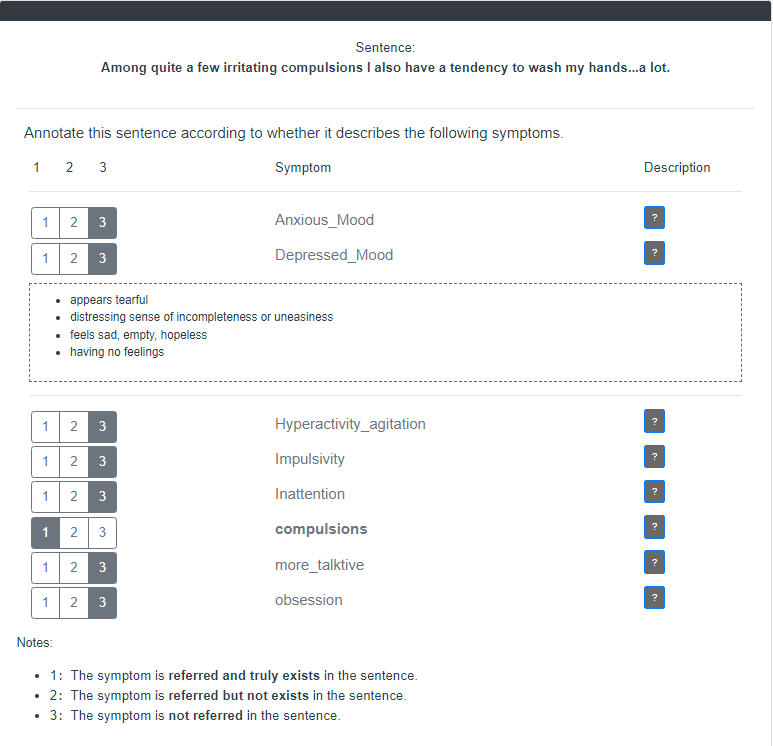}
    \caption{Screenshot of the annotation interface.}
    \label{fig:anno_interface}
\end{figure*}

\subsection{Candidate Selection Algorithm}
\label{apd:cand}

Previous attempts to establish symptom annotation dataset \citep{mowery2017understanding} have shown that the distribution of different symptoms can be highly unbalanced, with many classes having too few positive samples to train a plausible classifier. On the other hand, we found in our preliminary annotations that it's hard to achieve high agreement on atypical expressions of symptoms. Therefore, to encourage a more balanced symptom distribution, while retrieving the most typical symptom posts for high annotation agreement, we designed a priority queue based algorithm for the selection of candidate posts with embedding similarity (\S \ref{sec:data_retrieval}):
\begin{enumerate}
    \item For the annotation of a disease $d$, we set up a queue with max capacity of 300 sentences for each symptom of $d$
    \item For each sentence $x$ in $d$'s related subreddits, and each symptom $s$ of $d$, we estimate the $x$'s relevance to $s$ with its max similarity with all sub-symptoms of $s$. We add $x$ into the queue of $s$, if it is not full, or $x$'s relevance is larger than the minimum one in $s$
    \item We aggregate the sentences from all queues, and perform deduplication with MinHash and Local Sensitive Hashing (LSH) \citep{leskovec2020mining} to eliminate identical or similar sentences. The remaining posts will be used as the annotation candidates for $d$.
\end{enumerate}

Since we only use sentences with the highest estimated relevance to symptoms, the selected posts tend to show more typical symptom expressions. The equal size of each symptom queue can further alleviate the class imbalance problem (see class distribution in Table \ref{tab:symp_count}) (but not totally eliminate it, because of the error of relevance approximation, the potential overlapping between queues and the inherent unbalanced distribution with in the original posts.)

\subsection{Status Annotation}

Initially, the symptom status is annotated on the basis of each symptom in a sentence. The reason is that a sentence can express multiple symptoms with different status. For example, in ``I don't have insomnia any more, but I'm still depressed.'', the status of the previous symptom is negative while that of the latter is positive. However, after annotation, we find that the negative status annotation for each symptom is scarce. We then consider merging all symptom-level status into sentence-level. The motivation are two-folds. First, we observed that most sentences with negative status share similar characteristics like having negation words or being a question. Moreover, we found that the conflicts between status annotations of different symptoms in the same sentence is rare, accounting for only 2.8\% of the sentences. We thus set the sentence-level status as negative if there is any negative status annotation in its relevant symptoms. If not specified, the PsySym discussed in this paper refers to the version with sentence-level status labels.

\subsection{Quality Control}

\begin{table}[ht]
    \small
    \centering
    \begin{tabular}{cc}
    \hline
    Disease & \# of sentences   \\
    \hline
    ADHD & 528\\
    Anxiety & 2822\\
    Bipolar Disorder & 1131 \\
    Depression & 1433\\
    Eating Disorder & 907\\
    OCD & 449\\
    PTSD & 1284\\
    \hline
    \end{tabular}
    \caption{Annotated number of sentences for each disease.}
    \label{tab:disease_count}
\end{table}

\begin{table}[ht]
    \small
    \centering
    \begin{tabular}{m{4.8cm}m{0.9cm}m{0.9cm}}
    \hline
    Symptom & \#Positive & $\kappa$ \\
    \hline
    Anxious Mood 	&	1790	&	0.736 	\\
    Autonomic symptoms	&	571	&	0.786 	\\
    Cardiovascular symptoms	&	506	&	0.937 	\\
    Catatonic behavior	&	231	&	0.668 	\\
    Decreased energy tiredness fatigue	&	250	&	0.728 	\\
    Depressed Mood	&	846	&	0.637 	\\
    Gastrointestinal symptoms	&	340	&	0.943 	\\
    Genitourinary symptoms	&	260	&	0.951 	\\
    Hyperactivity agitation	&	277	&	0.603 	\\
    Impulsivity	&	157	&	0.776 	\\
    Inattention	&	401	&	0.801 	\\
    Indecisiveness	&	151	&	0.779 	\\
    Respiratory symptoms	&	464	&	0.916 	\\
    Suicidal ideas	&	287	&	0.915 	\\
    Worthlessness and guilty	&	291	&	0.614 	\\
    avoidance of stimuli	&	78	&	0.495 	\\
    compensatory behaviors to prevent weight gain	&	423	&	0.869 	\\
    compulsions	&	168	&	0.767 	\\
    diminished emotional expression	&	179	&	0.547 	\\
    do things easily get painful consequences	&	316	&	0.804 	\\
    drastical shift in mood and energy	&	425	&	0.889 	\\
    fear about social situations	&	415	&	0.938 	\\
    fear of gaining weight	&	257	&	0.747 	\\
    fears of being negatively evaluated	&	127	&	0.606 	\\
    flight of ideas	&	237	&	0.810 	\\
    intrusion symptoms	&	260	&	0.720 	\\
    loss of interest or motivation	&	181	&	0.676 	\\
    more talktive	&	165	&	0.647 	\\
    obsession	&	433	&	0.748 	\\
    panic fear	&	419	&	0.808 	\\
    pessimism	&	332	&	0.707 	\\
    poor memory	&	264	&	0.854 	\\
    sleep disturbance	&	320	&	0.852 	\\
    somatic muscle	&	428	&	0.879 	\\
    somatic symptoms others	&	347	&	0.723 	\\
    somatic symptoms sensory	&	258	&	0.777 	\\
    weight and appetite change	&	486	&	0.794 	\\
    Anger Irritability	&	427	&	0.841 	\\
    \hline
    \end{tabular}
    \caption{Number of positive samples (multiple annotations of a sentence aggregated) and Fleiss's $\kappa$ of symptoms}
    \label{tab:symp_count}
\end{table}

\begin{table}[h]
    \small
    \centering
    \begin{tabular}{lc}
    \hline
    Disease          & \# Users \\
    \hline
    Depression       & 3105         \\
    Anxiety          & 2239         \\
    Autism           & 716         \\
    ADHD             & 2374         \\
    Bipolar Disorder & 1366         \\
    OCD              & 753          \\
    PTSD             & 391          \\
    Schizophrenia    & 345          \\
    Eating Disorder  & 138          \\
    \hline
    \end{tabular}
    \caption{Number of users suffering from each disease in the disease detection dataset.}
    \label{tab:disease_detect_count}
\end{table}

\paragraph{Score Calculation} One core component of our quality control measures is the annotation quality score. Given the annotations and references (judgment from the authors), the quality score is calculated as the symptom-level $F_{\beta}$ score between them. The $\beta$ is set to 2 to put more emphasis on recall. 

The score is used in several phases of the annotation \ref{sec:data_annotation}. In \textit{Screening Tests}, the annotators need to achieve above 75 score to be qualified to conduct further annotation. We will also calculate the score in \textit{Sampling Inspection}, and reject all annotations in the checking batch if the score is below 60. To motivate accurate annotation, the volunteers will be rewarded with higher subsidy for high scores achieved according to our inspection.

\paragraph{Annotation Interface} The annotation interface is a customized webpage with many designs to promote efficient and accurate annotations. First, the annotator will be enter a disease description page, where all symptoms and their descriptions are shown for annotators to get familiar with. They will then enter the annotation page (Figure \ref{fig:anno_interface}). All symptom names are shown in list below the sentence to annotate. Annotators can click on the question mark beside the symptom to expand its descriptions, which can serve as a quick reminder. We will also make a symptom bold, if the sentence hit any keywords in the symptom descriptions, which can help the annotators quickly focus on likely classes, and also prevent them from missing annotations in case the annotators are unaware of some items in the symptom descriptions.

\subsection{Detailed Data Statistics}
\label{apd:stats}

We show the number of annotated sentences for each disease in Table \ref{tab:disease_count}, where diseases with more symptoms will have more annotations. We also show the number of annotations and the inter-annotator agreement by each symptom in Table \ref{tab:symp_count}. It can be seen that almost all symptom classes received reasonable amount of annotations, and have high agreement, indicating the effectiveness of proposed annotation framework to guarantee the quality of PsySym.

For the disease detection dataset (\S \ref{sec:data_disease}), we also provide the number of users suffering from each disease in Table \ref{tab:disease_detect_count}. The distribution of the 9 diseases are similar to SMHD \citep{cohan2018smhd}.

\section{Detailed Experimental Settings}
\label{apd:settings}

\begin{table*}[t]
    \centering
    \begin{tabular}{lcccc}
    \hline
    Disease           & TF-IDF & BERT  & Symp  & Symp (Reweighting) \\ \hline
    Depression        & 68.95  & 71.58 & 66.67 & 69.88              \\
    Anxiety           & 66.48  & 71.08 & 71.82 & 71.39              \\
    ADHD              & 59.55  & 60.05 & 60.14 & 60.45              \\
    Bipolar Disorder  & 66.67  & 43.56 & 67.76 & 68.82              \\
    OCD               & 26.09  & 44.83 & 48.33 & 56.52              \\
    PTSD              & 30.43  & 27.45 & 48.48 & 45.9               \\
    Eating Disorder   & 11.11  & 37.04 & 52.17 & 46.15              \\ \hline
    Autism            & 30.23  & 49.59 & 35.64 & 37.04              \\
    Schizophrenia     & 34.04  & 57.97 & 48.15 & 57.63              \\ \hline
    Mean (7 Diseases) & 47.04  & 50.80 & 59.34 & \textbf{59.87}     \\
    Mean (9 Diseases) & 43.73  & 51.46 & 55.46 & \textbf{57.09}     \\ \hline
    \end{tabular}
    \caption{MDD results by disease. We distinguish the 7 disease types included in PsySym (upper) and the remaining two (lower), and calculate the average performance with 7 and 9 diseases.}
    \label{tab:mdd_by_disease}
\end{table*}

For all models, we empirically set hyperparameters following existing implementations and previous works without fine-tuning them for optimized performance. Specifically, we use a learning rate (lr) of 3e-4, max sequence length of 64 for BERT and MBERT models used for symptom relevance judgment and status inference. For the CNN model used in mental disease detection, the structure is identical to that of \citet{nguyen2022improving}, the lr is 0.01 when using symptom features only and 0.003 when using BERT embeddings or the combined features. The posting list will be truncated to preserve the earliest 256 posts at most. We also employ early-stopping with a patience of 4 epochs according to validation performance to prevent overfitting. For efficiency of inference on the MDD dataset with large amount of posts, we use bert-tiny \footnote{\url{https://huggingface.co/prajjwal1/bert-tiny}} models trained on PsySym to extract the symptom and status features, whose performance is close to the best performing MBERT model (AUC=98.80, F1=62.28 for relevance judgement with control posts and MAE=0.1509 for status inference). For the multi-task learning of multiple symptoms, the losses of each classes are the simple arithmetic mean of them without weighting. 

For each post, we conduct several preprocessing steps. First, they are split into sentences with \textit{blingfire}\footnote{\url{https://github.com/microsoft/BlingFire}}. We will then eliminate sentences like ``[Removed]''. We also use regular expressions to detect the hyperlink format like ``[anchor text](web url)'', and transform it into only anchor text.

\section{Detailed Disease Detection Results}
\label{apd:mdd_results}

We show the MDD performance of different methods on disease level in Table \ref{tab:mdd_by_disease}. It can be seen that the proposed symptom-based methods perform slightly worse than BERT on the 2 diseases not included in PsySym, which is reasonable. When comparing only the 7 PsySym diseases the advantage of Symp over BERT is even more significant, which further demonstrate the strength of the proposed method. We also observe a more significant improvement on diseases with fewer positive samples like OCD, PTSD and Eating Disorder, suggesting the usefulness of symptom knowledge for the MDD in low-resource scenarios.

\section{Evaluation of Candidate Retrieval Strategy}
\label{apd:exp_retrieve}

In this section, we show the benefits of the proposed embedding-based candidate retrieval strategy with quantitative experimental results, compared with the keyword/pattern matching methods commonly used in previous works \citep{mowery2017understanding,yadav2020identifying}. Specifically, we test the effectiveness of different strategies in retrieving the positive samples among all annotated sentences of PsySym, and report the average \textit{precision} and \textit{recall} across 38 symptoms. A low \textit{recall} would harm the diversity of the annotated corpus, as we would be unable to annotated on sentences supposed to be symptom relevant, while a low \textit{precision} will lead to low annotation efficiency, since we will have to read through more posts in order to find those actually convey the symptom. 

The compared methods are: 

\paragraph{MeSH} The Medical Subject Headings (MeSH) thesaurus is a large-scale vocabulary of medical terms produced by National Library of Medicine (NLM)\footnote{\url{https://www.nlm.nih.gov/mesh/meshhome.html}}. We use the entity linker in scispacy \citep{neumann2019scispacy} to detect all MeSH terms (with alias) related to somatic symptoms and mental status in the sentence\footnote{Specifically, we detect terms under the class of Signs and Symptoms [C23.888], Emotions [F01.470] or Behavioral Symptoms [F01.145.126]}, and sentence with any single matched term will be retrieved. We tried to estimate its recall upper bound by greedily including all symptom terms so that some of them may be beyond the scope of mental health, and we thus don't report its precision. 
\paragraph{LIWC (negemo)} Linguistic Inquiry and Word Count (LIWC) \citep{pennebaker2001linguistic} is a categorized vocabulary that can provide useful dimensions to analyze a person's thoughts, feelings, personality from language use. It has been shown in many works that some dimensions of LIWC are closely related to mental disorders \citet{shen2017depression,cohan2018smhd}, especially the Negative Emotion (negemo) words that include sadness, anxiety and anger related words, which are also symptoms in our KG. To leverage LIWC (negemo), we simply retrieve all sentences that contain any single negative emotion words. 

\paragraph{SBERT} This is our proposed method for candidate retrieval (\S \ref{sec:data_retrieval}), which leverages embedding similarity instead of keyword matching used in the previous two methods. We study two variants of this method. \textbf{SBERT (manual only)} is our first attempt that only uses the symptom description collected from DSM-5 and clinical questionnaires. \textbf{SBERT (manual+post)} additionally incorporates representative posts of a symptom for some symptom classes that we found to have poor retrieval results with the previous methods. To get a binary retrieval decision for the calculation of precision and Recall, we use 0.5 to threshold on the calculated cosine similarity. This does not totally reflect the actual setting in our data collection, but allows a direct comparison with previous methods.

\begin{table}[h]
    \centering
    \begin{tabular}{lcc}
    \hline
    Method                 & Precision   & Recall    \\ 
    \hline
    MeSH                   & / & 37.12 \\
    LIWC (negemo)          & 1.23 & 67.50 \\
    \hline
    SBERT (manual only)    & \textbf{50.36} & 66.12 \\
    SBERT (manual+post)    & 48.89 & \textbf{77.09} \\
    \hline
    \end{tabular}
    \caption{Candidate Retrieval performance on the annotated sentences of PsySym.}
    \label{tab:cand_retrieve}
\end{table}

We can see from Table \ref{tab:cand_retrieve} that the recall of \textit{MeSH} is not high despite our greedy inclusion of matching terms. This suggests can keyword matching with professional terms (even with alias in MeSH) failed to identify the diverse expressions of symptoms on social media potentially due to its figurative language \citep{yadav2020identifying}. \textit{LIWC (negemo)} received a relatively high recall at the expense of precision. The possible reason is that the negative emotion words are too broad and does not target certain symptom. \textit{SBERT (manual only)} is enough to achieve both satisfying precision and recall. The symptom-specific descriptions can precisely detect candidates for each symptom, and embedding based retrieval can overcome the limitation of exact word matching methods, be tolerant to misspelling and synonyms, and recall sentences expressing the semantics of a symptom but with no specific keywords. \textit{SBERT (manual+post)} can further improve the recall while almost preserving the precision. This indicates that the inclusion of posts can alleviate the mismatching between manual and candidate posts in the language style and the perspective (observation versus self expression). 
Note that our method does not achieve 100\% recall at the dataset because the evaluation is conducted at a pre-symptom basis, and a sentence related to symptom A can also be retrieved with the descriptions of symptom B, which is not counted in the recall of symptom A.

\end{document}